\documentclass[sigconf]{acmart}

\usepackage{url}
\usepackage{graphicx}
\usepackage{amsmath}
\usepackage{amsthm}
\usepackage{booktabs}
\usepackage{algorithm}
\usepackage{algorithmic}
\usepackage{booktabs}
\usepackage{multirow}
\usepackage[utf8]{inputenc}

\AtBeginDocument{%
  \providecommand\BibTeX{{%
    \normalfont B\kern-0.5em{\scshape i\kern-0.25em b}\kern-0.8em\TeX}}}

\begin{document}


\title{Recent Advances on Federated Learning: A Systematic Survey}

\author{Bingyan Liu}
\email{bingyanliu@bupt.edu.cn}
\affiliation{%
  \institution{Beijing University of Posts and Telecommunications}
  \city{Beijing}
  \country{China}
  \postcode{100871}
}

\author{Nuoyan Lv}
\email{lvnuoyan@bupt.edu.cn}
\affiliation{%
  \institution{Beijing University of Posts and Telecommunications}
  \city{Beijing}
  \country{China}
  \postcode{100871}
}

\author{Yuanchun Guo}
\email{gyc2001@bupt.edu.cn}
\affiliation{%
  \institution{Beijing University of Posts and Telecommunications}
  \city{Beijing}
  \country{China}
  \postcode{100871}
}

\author{Yawen Li}
\email{warmly0716@126.com}
\affiliation{%
  \institution{Beijing University of Posts and Telecommunications}
  \city{Beijing}
  \country{China}
  \postcode{100871}
}







\renewcommand{\shortauthors}{Liu et al.}

\begin{abstract}
Federated learning has emerged as an effective paradigm to achieve privacy-preserving collaborative learning among different parties. Compared to traditional centralized learning that requires collecting data from each party, in federated learning, only the locally trained models or computed gradients are exchanged, without exposing any data information. As a result, it is able to protect privacy to some extent. In recent years, federated learning has become more and more prevalent and there have been many surveys for summarizing related methods in this hot research topic. However, most of them focus on a specific perspective or lack the latest research progress. In this paper, we provide a systematic survey on federated learning, aiming to review the recent advanced federated methods and applications from different aspects. Specifically, this paper includes four major contributions. First, we present a new taxonomy of federated learning in terms of the pipeline and challenges in federated scenarios. Second, we summarize federated learning methods into several categories and briefly introduce the state-of-the-art methods under these categories. Third, we overview some prevalent federated learning frameworks and introduce their features. Finally, some potential deficiencies of current methods and several future directions are discussed.

\end{abstract}



\keywords{Decentralized AI, Federated Learning, Neural Networks, Survey}


\maketitle

\section{Introduction}

Over the past few years, deep neural networks (DNNs) have received a lot of attention due to their remarkable performance on various tasks such as Computer Vision (CV) \cite{krizhevsky2012imagenet,simonyan2014very,liu2019wealthadapt, liu2021transtailor}, Natural Language Processing (NLP) \cite{devlin2018bert,vaswani2017attention,DBLP:conf/acl/XuZGZL20}, Recommendation Systems (RS) \cite{10.1145/3485447.3511997,chen2020efficient,chen2020jointly} and Data Mining (DM) \cite{liu2020pmc,li2021modeldiff, shao2014parallel}. However, the superiority of DNNs depends on the support of big data, which is hard to access in a certain party considering the limitation of the storage space and the difficulty of data collection. Gathering data from different parties to a central server for training is a direct solution to the issue. Nevertheless, data in each party may be sensitive or include some user privacy information. For example, medical images in a hospital are prohibited from outsourcing due to their privacy property. Besides, policies such as  General
Data Protection Regulation (GDPR) \cite{albrecht2016gdpr} also highlight the importance of protecting privacy when sharing information among different organizations. Thus, how to aggregate the data knowledge from different parties while ensuring privacy is an important and practical problem in real-world scenarios.

Federated learning (FL) \cite{mcmahan2017communication}, which enables multiple parties to collaboratively train a DNN with the help of a central server, can be regarded as an effective solution to the aforementioned problem. Different from the traditional centralized learning that needs to collect data from each party, in FL, data do not need to upload for a joint training. Instead, the local trained models are exchanged with a central server, which are used to aggregate the knowledge from all of the uploaded models and then distribute the global model to each party. As a result, each party is able to benefit from other parties, improving the model accuracy. In recent years, there have been 
many applications based on FL in practice, such as loan status prediction, health situation assessment, and next-word prediction \cite{hard2018federated,yang2018applied,yang2019federated}.

\begin{figure*}[t]
\centering
\includegraphics[width=2.1\columnwidth]{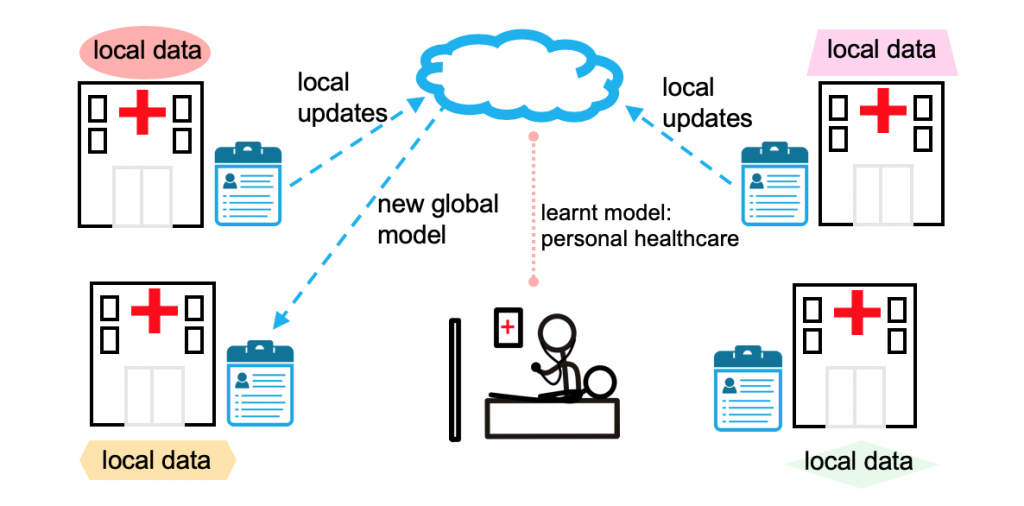} 
\caption{An example of the FL pipeline \cite{li2020federated}.}
\label{fig:pipeline}
\end{figure*}

We take Fig. \ref{fig:pipeline} as an example to illustrate a typical FL pipeline. First, each hospital (party) trains the local model distributed from a central cloud. The training process is usually implemented based on SGD with local data and then generates corresponding local updates. Second, the local updates rather than local data are transferred to the cloud, where the updates are sampled in terms of some heuristic rules to ensure the overhead and some aggregation algorithms (e.g., FedAvg \cite{mcmahan2017communication}) are conducted to achieve effective knowledge integration. In this way, the cloud can get an improved new global model and distributes it to each hospital for further tuning. These steps may repeat several times until the healthcare service can be satisfied (e.g., the accuracy of the learned model is acceptable for practical deployment). 

\begin{figure}[t]
\centering
\includegraphics[width=1\columnwidth]{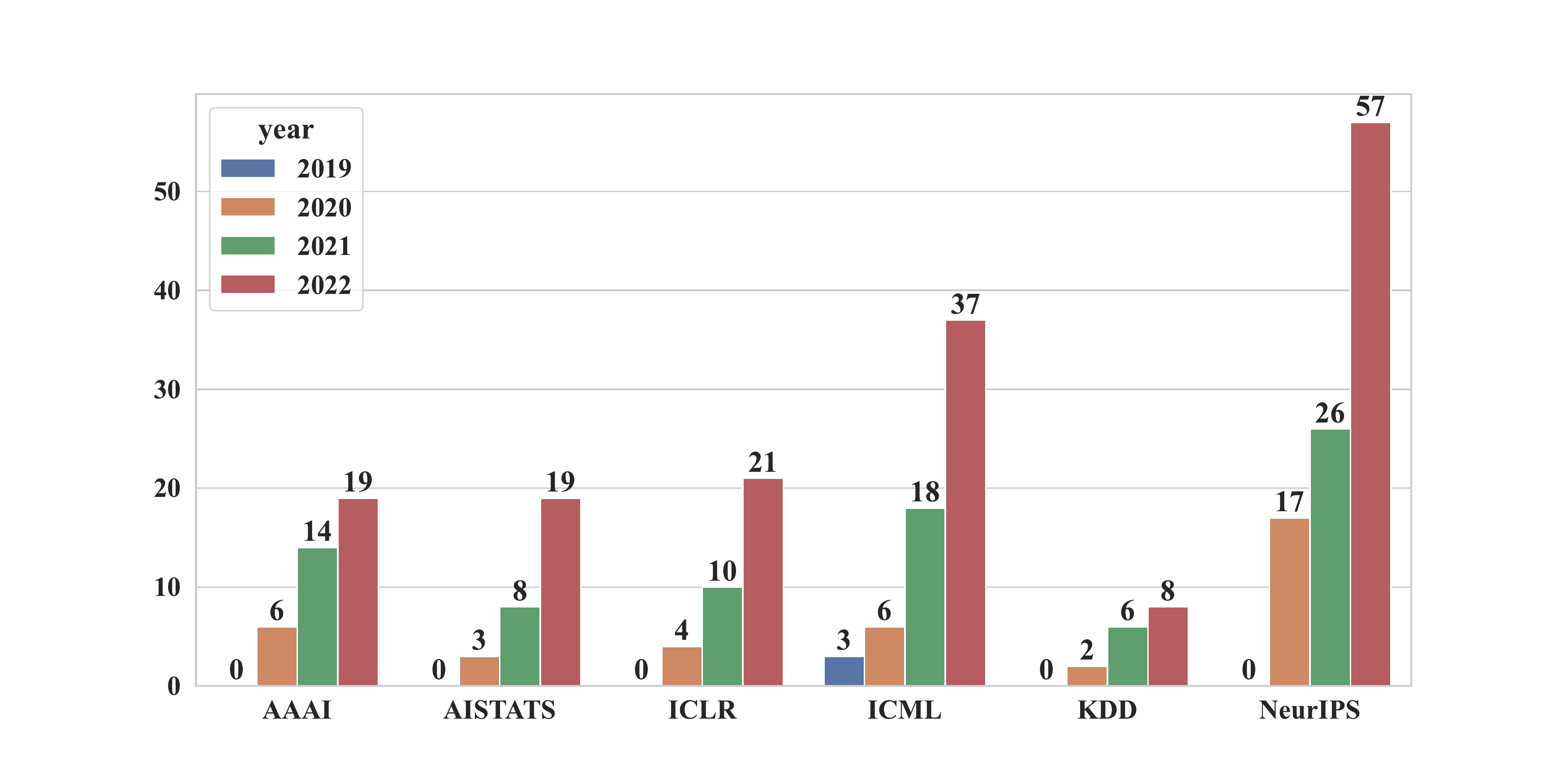} 
\caption{The number of pulished FL papers in top-tie conference from 2019-2022.}
\label{fig:fl_trend}
\end{figure}

There have been other surveys on FL over the past few years. For instance, Li \textit{et al.} \cite{li2021survey} summarized related FL methods from the system perspective, where the authors provided the definition of federated learning systems and analyzed the system
components. Lim \textit{et al.} \cite{lim2020federated} focused on the FL application in mobile edge networks. Lyu \textit{et al.} \cite{lyu2020threats} paid more attention to the security and privacy issues existed in current FL schemes. However, these surveys only review a specific aspect of federated learning, failing to give readers a comprehensive understanding on FL. Towards the general FL overviews, most of them are out of date and cannot catch the latest trend in FL research. For example, Yang \textit{et al.} \cite{yang2019federated} divided FL methods into three categories (i.e., horizontal federated learning, vertical federated learning and federated transfer learning) and described their features respectively. Kairouz \textit{et al.} \cite{kairouz2021advances} gave a comprehensive introduction of federated learning theory and application. Notice that both of the surveys mainly cited papers published before 2020, which is impossible to track the latest research progress on FL considering the rapid development in this field. As shown in Fig. \ref{fig:fl_trend}, we can clearly see that the number of accepted FL papers in top-tier conferences increases dramatically after 2020, which calls for a timely survey to summarize the advances in the FL community. Besides, the rapid update of FL frameworks also requires us to highlight their latest features.

In this paper, we attempt to provide a systematic survey on federated learning, targeting at reviewing the recent advanced federated methods and applications from different aspects. Specifically, the key contributions of this survey are as follows: (1) we present a new taxonomy based on the federated learning pipeline and challenges, which includes four typical aspects: \textit{aggregation optimization, heterogeneity, privacy protection, fairness}. We will give detailed explanation in the following sections. (2) we summarize different federated learning methods into the proposed categories and briefly describe the state-of-the-art methods under these categories. (3) we overview the latest federated learning frameworks and introduce their features. (4) we discuss some potential deficiencies of current methods and several future directions.

The remainder of this survey is structured as follows. In Section \ref{sec:preliminaries}, we first introduce preliminaries of federated learning. In Section \ref{sec:Approaches}, we propose the taxonomy of federated learning according to different aspects, in which various federated learning approaches are discussed and categorized. Then, in Section \ref{sec:framework}, we introduce some prevalent frameworks to show the practical deployment of federated learning. Finally, Section \ref{sec:discussion} and Section \ref{sec:conclusion} discuss the future work and concludes this paper.

\section{Preliminaries}
\label{sec:preliminaries}

\subsection{Problem formulation}
In this section, we first introduce some notations and symbols used in this survey to formally define federated learning. In general, there are two ends participated in the round of federated learning: \textit{client end} and \textit{server end}. The client end holds a series of local private data $\mathcal{D}=\{\mathcal{D}_1,\mathcal{D}_2,...,\mathcal{D}_N\}$, which are then used to train the model in each client and generate local models $\mathcal{M}=\{M_1,M_2,...,M_N\}$. Here $N$ denotes the number of clients. After the local training process, the local models $\mathcal{M}$, rather than the data $\mathcal{D}$, are uploaded to the server end, where aggregation algorithms are implemented to obtain a global model $M_{global}$. The process can be defined as
\begin{align}
\label{equation:agg}
   M_{global}=AGG(M_1,M_2,...,M_N),
\end{align}
where $AGG$ represents the aggregation algorithms. In this way, we finish one round of federated learning and distribute the global model to each client side for further local training. The concrete number of round is usually determined by the model performance (i.e., we stop the process until the model can achieve desirable accuracy). In addition, to provide a more rigorous privacy protection, each client may enforce some encryption techniques to the models before uploading them. Differential privacy (DP) \cite{dwork2008differential} and homomorphic encryption (HE) \cite{gentry2009fully} are widely used to conduct such protection. 

Based on the aforementioned statement, we can see that the performance of federated learning largely depends on the aggregation algorithm in the server end. Formally, the goal of federated learning is to optimize the following objective function
\begin{align}
\label{equation:goal}
   \min_{w} \ L(w), \ where \ L(w)=\sum_{i=1}^N f_i L_i(w),
\end{align}
where $w$ is the weights of DNNs, $L(w)$ is the global loss function and $L_i(w)$ is the local loss function in the $i_{th}$ client. $f_i$ represents the importance of the $i_{th}$ client and $\sum_{k=1}^N=1$. In federated learning, the aggregation algorithm determines the value allocation for $f_i$. Many research papers that try to improve the accuracy performance of federated learning are focused on this aspect.

\subsection{Key challenges}
Different from traditional centralized learning or distributed learning, federated learning faces the following key challenges:
\begin{itemize}
    \item \textbf{Heterogeneity problem.}
    In federated learning, the heterogeneity comes from three aspects:(1) Data heterogeneity. Considering that each participator collects data from its local end,  the overall data distribution inevitably conforms to the non-independent identically distribution (non-iid) situation. For example, the same object image collected from different environments, or the same activity coming from different people, can lead to different data distributions, which will further affect the performance of federated aggregation \cite{zhao2018federated}. (2) Model heterogeneity. In real-world scenarios, it is hard to limit the federated clients to use an identical model architecture. Instead, each client may prefer a distinctive model architecture for improved task performance. Therefore, how to aggregate these heterogeneous models is challenging in practical federated learning conditions. (3) System heterogeneity. Because of the variability in hardware, different parties may have different storage space, computation power, and communication capabilities. As a result, the server end needs to decide whether to wait for all parties to upload their models for better accuracy or remove stragglers (i.e., the parties with weak hardware performance) for accelerating the federation process.
     
      \item \textbf{Privacy leakage.}
      The key idea of federated learning is to achieve collaborative learning in a privacy-preserving manner, which differs from the traditional paradigm that exchanges data or other sensitive information. Keeping data in the local end and transferring corresponding models is the original privacy protection design in federated learning. However, the parameters of the uploaded models may also be exploited by attackers to infer the user privacy information \cite{zhu2019deep}. So we require more rigorous encryption or obfuscation methods to ensure privacy.
      
       \item \textbf{Unfairness.} 
       In traditional centralized learning or distributed learning, the unfairness problem does not exist since the participants belong to a same organization. However, the participants in federated learning come from various parties with different data resources. According to a previous work \cite{dwork2012fairness}, if individuals with similar preferences and characteristics receive substantially different outcomes, then we say that the model violates individual fairness. Thus, it is necessary to generate federated models that go beyond average accuracy to further consider the fairness performance.
\end{itemize}

\section{Approaches of federated learning}
\label{sec:Approaches}
In this section, we first present a taxonomy of federated learning and allocate different federated approaches into different categories according to the taxonomy. Then for each category, we describe in detail how various methods achieve their goal.


\begin{figure*}[t]
\centering
\includegraphics[width=2.1
\columnwidth]{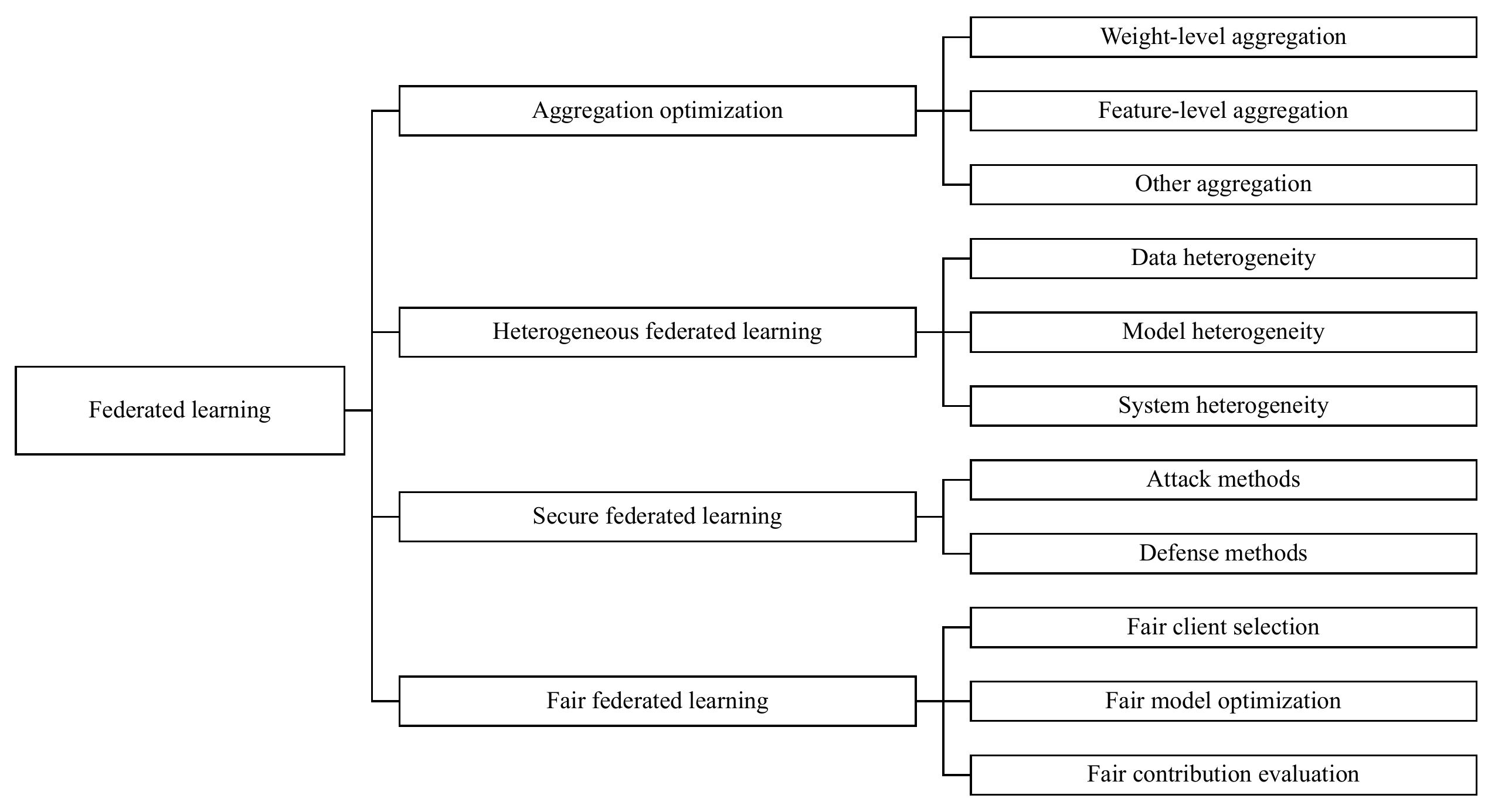} 
\caption{Our taxonomy of different federated learning methods.}
\label{fig:floverview}
\end{figure*}

\subsection{Taxonomy}
In this survey, we propose a new taxonomy to classify the existing federated learning methods (Fig. \ref{fig:floverview}).
Our taxonomy is motivated by the pipeline and challenges in federated learning. As stated in the previous section, the key step in the federated learning pipeline is the aggregation algorithm and the key challenges come from three different aspects. Therefore, in our taxonomy, federated learning approaches can be summarized into four cases: aggregation optimization, heterogeneous federated learning, secure federated learning and fair federated learning.
\begin{itemize}
    \item \textbf{Aggregation optimization.}
    Considering that the number of participants in a federated learning system is usually large, it is essential to implement an effective aggregation optimization for outputting a better global model compared to the ones with local training. This survey investigates various aggregation methods such as FedAvg \cite{mcmahan2017communication,nagalapatti2021game,zhao2018federated}, FedMA \cite{wang2020federated} and FedProx \cite{li2020fedprox}, with a focus on how to combine local models into an improved global model. 
    
    \item \textbf{Heterogeneous federated learning.}
    In real-world scenarios, federated clients may come from different environments or equip with various hardware, leading to the heterogeneity problem. In the following sections, we respectively explore how related research efforts address the issue of data heterogeneity, model heterogeneity and system heterogeneity. In particular, techniques such as meta-learning \cite{fallah2020personalized,acar2021debiasing,jiang2019improving,khodak2019adaptive,zheng2021federated,chen2018federated}, multi-task learning \cite{smith2017federated,vanhaesebrouck2017decentralized,corinzia2019variational,zantedeschi2020fully,huang2021personalized,li2021ditto,marfoq2021federated,he2021spreadgnn,zhou2022efficient,chen2022fedmsplit}, transfer learning \cite{wang2019federated,yu2020salvaging,peterson2019private,ozkara2021quped} and clustering \cite{sattler2020clustered,ghosh2020efficient,ghosh2019robust,zhang2020personalized,ruan2022fedsoft,lubana2022orchestra} are incorporated to achieve our goal.

    \item \textbf{Secure federated learning.}
    Although traditional federated learning has attempted to protect data privacy by only exchanging parameters of the local trained models, malicious attackers can still design some scheme to infer the properties of raw data. In our survey, we first summarize a series of attacks targeting federated learning, where we describe how backdoor attack \cite{bagdasaryan2020backdoor,sun2019can,wang2020attack,xie2019dba,zhang2022neurotoxin,xie2021crfl,ozdayi2021defending}, gradients attack \cite{zhu2019deep,lam2021gradient,hitaj2017deep,zhao2020idlg,yin2021see,li2022auditing,zhu2020r,geiping2020inverting} and poison attack \cite{bhagoji2019analyzing,sun2021fl,panda2022sparsefed,wu2022fedattack} are applied to compromise federated learning. Then we introduce how to combine federated learning, differential privacy (DP) \cite{wei2020federated,geyer2017differentially,mcmahan2018learning,kairouz2021distributed,agarwal2021skellam,zhang2022understanding,zheng2021federated,wang2020federated,girgis2021shuffled}, homomorphic encryption (HE) \cite{hardy2017private,zhang2020batchcrypt}, trusted execution environment (TEE) \cite{mo2021ppfl,mo2020darknetz} and other algorithms \cite{baruch2019little,xie2021crfl,huang2021evaluating,tang2022virtual} to defend aforementioned attacks.
    
    \item \textbf{Fair federated learning.}
     During federated learning, it is possible that the performance of the global model varies significantly across the devices, resulting in the fairness problem. This survey reviews literature about how to ensure fair federated learning, such as designing minimax optimization strategies \cite{sharma2022federated,tarzanagh2022fednest} and sample reweighting approaches \cite{zhao2022dynamic,enthoven2021fidel}.
\end{itemize}


\subsection{Aggregation optimization}
The goal of aggregation optimization is to improve the performance of the final global model, which is the core output in federated learning. There have been a large number of aggregation algorithms proposed to combine these local models to a better global one. In the following parts, we will describe in detail how different types of aggregation methods work.

\subsubsection{Weight-level aggregation}

A typical and prevalent weight-level aggregation method called FedAvg \cite{mcmahan2017communication} is mostly adopted by developers. The key idea of FedAvg is to aggregate these local models in a coordinate-based weight averaging manner, which can be denoted as
\begin{align}
\label{equation:descriptor}
    W_g^r=\frac{1}{N} \sum_{k=1}^{N} w_k^r,
\end{align}
where N is the number of federated clients. $w_k$ denotes the weight parameters of the $k_{th}$ client and $W_g^r$ is the final aggregated model at the $r_{th}$ round. Researchers have shown the remarkable performance of FedAvg on a variety of public datasets (e.g., MNIST \cite{lecun1998gradient} and CIFAR-10 \cite{krizhevsky2009learning}) and provided some theoretical analyses to prove why FedAvg works well \cite{li2019convergence}.

Despite being widely applied, FedAvg still suffers from the weight divergence problem \cite{zhao2018federated}: the weight in the same coordinates (e.g., same layer or same filter) may have a large mismatching due to the highly skewed data distribution in each distinctive client/party. Therefore, directly averaging them will degrade the accuracy of the generated global model. To solve the issue, researchers leverage a particular DNN principle, \textit{weight permutation invariance}, which has been mentioned and discussed by recent works \cite{wang2020federated, yurochkin2018probabilistic,yurochkin2019bayesian}. The key idea of this principle is that the weights in a DNN can be specially shuffled without incurring much accuracy drop. Concretely, suppose $l_j$ and $l_{j+1}$ are the weight of two continuous layers in a DNN model, where the output function can be denoted as 
\begin{align}
\label{equation:dnn_comput}
    O_{j+1}=l_{j+1} l_j  I,
\end{align}
where $I$ is the input and $O_{j+1}$ is the output of the ${j+1}_{th}$ layer. Note that for each weight matrix $l$, it can be further decomposed as follows
\begin{align}
\label{equation:permutation}
    l=l \mathbf{1}=l \Pi \Pi^{T},
\end{align}
where $\Pi$ represents the permutation matrix. In terms of this equation, we can transform Eq. \ref{equation:dnn_comput} to the following form
\begin{align}
\label{equation:permutation_invar}
    O_{j+1}=(l_{j+1} \Pi_{j+1}\Pi^{T}_{j+1}) l_j  I=(l_{j+1} \Pi_{j+1}) (\Pi^{T}_{j+1} l_j)  I,
\end{align}
Based on Eq. \ref{equation:permutation_invar}, we can clearly see that the original layer weight can be losslessly transformed with a pair of well-designed permutation matrices, which we call it \textit{weight permutation invariance}. 

In federated learning, traditional aggregation methods fuse local models according to their weight location, which may be sub-optimal since the \textit{weight permutation invariance} principle indicates that we can change the weight value in a specific location while ensuring the same performance. Thus, the location-based aggregation cannot achieve accurate knowledge fusion, leading to the weight mismatching problem.

\begin{figure}[t]
\centering
\includegraphics[width=1\columnwidth]{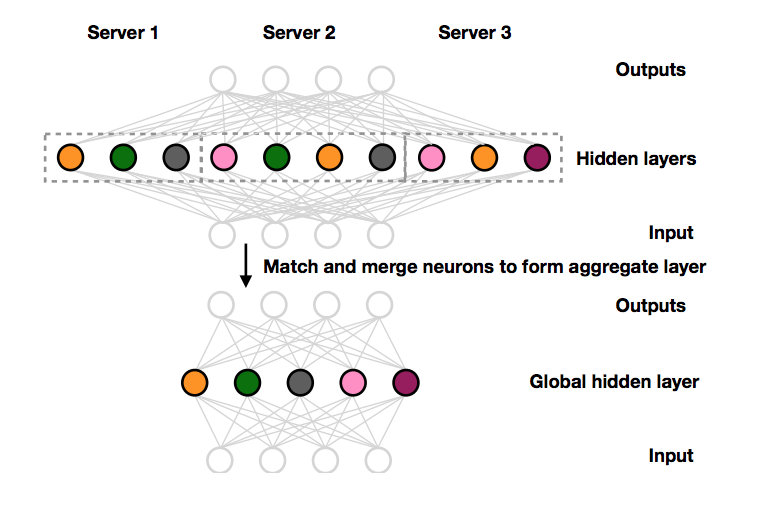} 
\caption{The illustration of PFNM \cite{yurochkin2019bayesian}.}
\label{fig:PFNM}
\end{figure}

To address this problem, a large number of federated optimation works attempt to achieve weight-level alignment. For example, Yurochkin \textit{et al.} \cite{yurochkin2019bayesian} developed Probabilistic Federated Neural Matching (PFNM). As shown in Fig. \ref{fig:PFNM}, the key idea is to identify subsets of
neurons in each local model that matches neurons in other local models and then  combine the matched neurons to an improved global model by leveraging Bayesian nonparametric machinery. For single-layer neural matching, they presented a Beta Bernoulli Process \cite{zhang2013information} based model of MLP weight parameters, where the corresponding neurons in the
output layer are used to convert the neurons in each batch and form a cost matrix. Then the matched neurons can be
aggregated to generate the final global model. For multilayer neural matching, they extended the single strategy by defining a generative
model of deep neural network weights from outputs back to inputs. In this way, they could adopt a greedy inference procedure that first infers the matching of the top layer and then proceeds down the layers of the model. 

Unfortunately, PFNM only performs well on simple architectures (e.g. fully connected feedforward networks). For more complex CNNs and LSTMs, it just receives minor improvements over location-based methods (e.g., FedAvg). To further achieve the weight alignment goal, Wang \textit{et al.} \cite{wang2020federated} proposed Federated Matched Averaging (FedMA) to effectively align advanced CNNs and LSTMs in a layer-wise manner. The key idea is to search for the best permutation matrices by addressing the following optimization problem
\begin{align}
\label{equation:permutation_opt}
    &\min _{\left\{\pi_{l i}^{j}\right\}} \sum_{i=1}^{L} \sum_{j, l} \min _{\theta_{i}} \pi_{l i}^{j} c\left(w_{j l}, \theta_{i}\right) 
    \\ &\text { s.t. } \sum_{i} \pi_{l i}^{j}=1 \forall j, l ; \sum_{l} \pi_{l i}^{j}=1 \forall i, j,
\end{align}
where $\theta_i$ is the $i_{th}$ neuron in the current global model, $w_{jl}$ is the  output weights processed by permutation matrix $\pi_{l i}^{j}$. $c()$ is the distance metric served as determining the similarity between neurons. To solve this optimization problem, unlike PFNM that used heuristic choices, FedMA addressed it by the Hungarian matching algorithm \cite{kuhn1955hungarian}.

\begin{figure}[t]
\centering
\includegraphics[width=1\columnwidth]{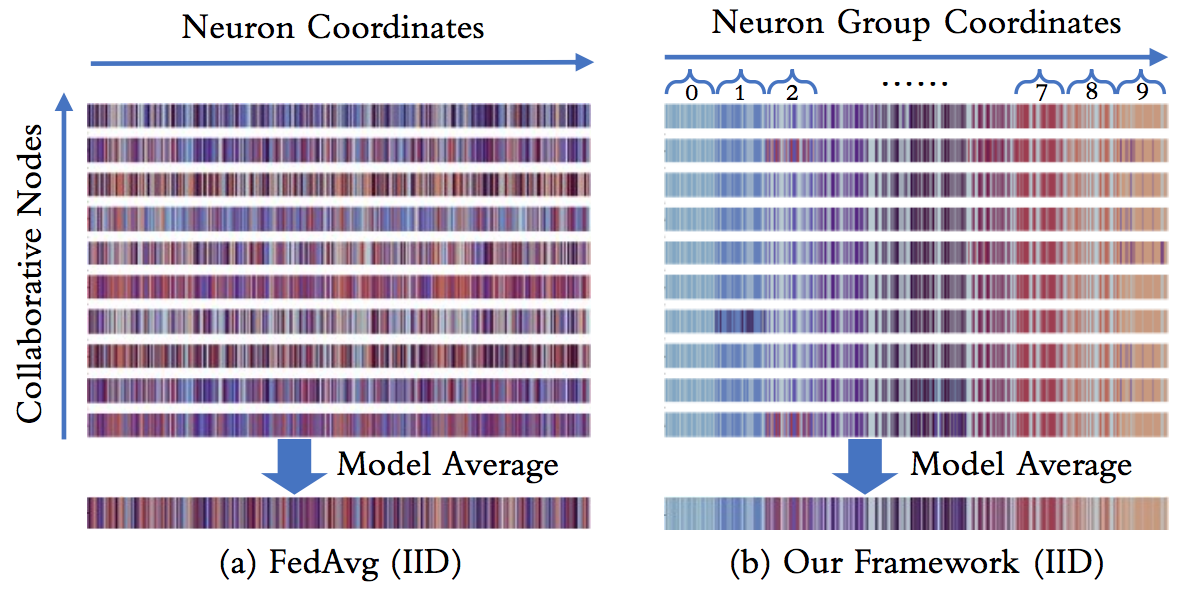} 
\caption{Comparison between FedAvg and $Fed^2$ \cite{yu2021fed2}.}
\label{fig:fvg-fed2}
\end{figure}

\begin{figure*}[t]
\centering
\includegraphics[width=2
\columnwidth]{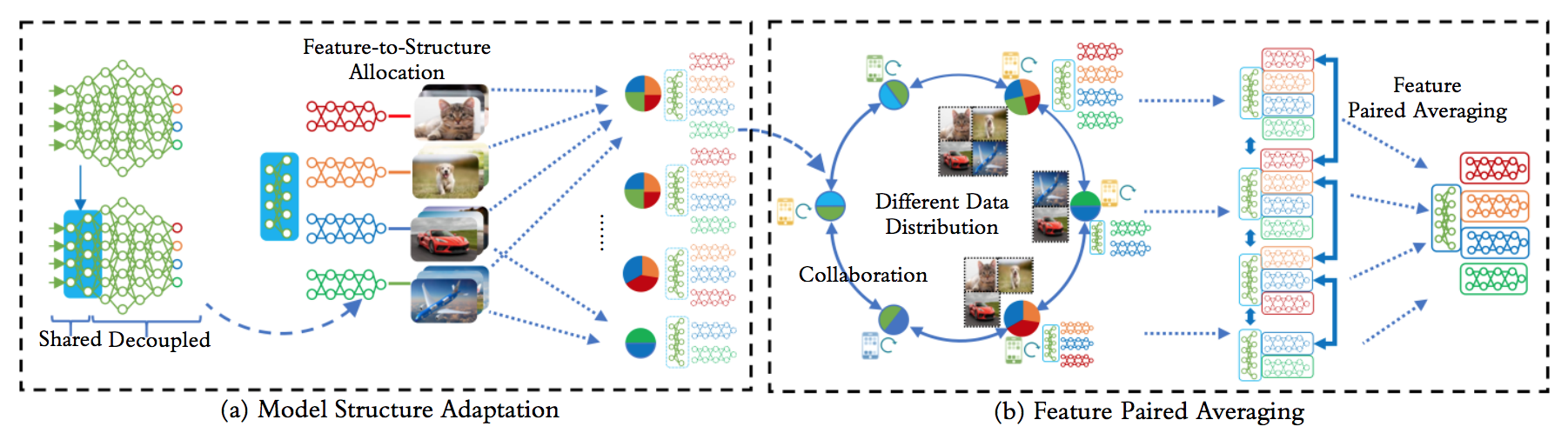} 
\caption{The illustration of $Fed^2$ \cite{yu2021fed2}.}
\label{fig:fed2}
\end{figure*}

\subsubsection{Feature-level aggregation}
Despite effectiveness, the performance of weight-level aggregation/alignment largely depends on the selection of distance metric, which may not fully reflect the inherent feature information embedded in the neurons. In addition, the computation cost of the matching process is significantly heavy. To address these limitations, Yu \textit{et al.} \cite{yu2021fed2} designed a feature-level alignment method, named $Fed^2$ which is composed of a feature-oriented structure adaptation and a model fusion algorithm. As shown in Fig. \ref{fig:fvg-fed2}, compared with traditional weight alignment, $Fed^2$ paid more attention to the neuron features and then aggregated the corresponding neurons. As a result, similar knowledge can be fused to achieve better performance. 

Concretely, the authors developed two schemes to accomplish feature-based federated learning. Fig. \ref{fig:fed2} shows the pipeline of the proposed $Fed^2$. The first scheme is \textit{model structure adaptation}, where $Fed^2$ takes advantage of the group-convolution technique to allocate and learn the distinctive neuron features. Next, a feature paired averaging policy is presented to aggregate different neurons according to the partitioned group features. In this way, $Fed^2$ enables more accurate feature alignment as well as avoiding the expensive distance-based optimization.

\subsubsection{Other aggregation}

The aforementioned works mainly focus on alignment, in fact, there are also many other literatures targeting federated aggregation. For example, Yin \textit{et al.} \cite{yin2018byzantine} proposed a robust aggregation method for distributed learning. In the beginning, this work mainly analyzed two robust distributed gradient descent (GD) algorithms, including the coordinate-wise median and the coordinate-wise trimmed mean. They proved statistical error rates for three kinds of population loss functions: strongly convex, non-strongly convex, and smooth non-convex. Furthermore, to reduce the communication cost, the authors designed a median-based distributed algorithm and demonstrate its effectiveness by extensive experiments. Chen \textit{et al.} \cite{chen2020distributed} further considered the federated learning scenario, and found that heterogeneous data in different nodes will harm the training convergence to some degree. Based on this observation, they developed a novel gradient correction mechanism that can perturb the local gradients with noise. The main advantage of the proposed scheme is that it offers a provable convergence guarantee even when data are non-iid.

Besides, Yurochkin \textit{et al.} \cite{yurochkin2019statistical} leveraged Bayesian nonparametrics to design a meta-model that can potentially capture the global structure through statistical parameter matching. The authors pointed out that their approach is  model-independent and is applicable to a wide range of model types. 
Chen \textit{et al.} \cite{chen2020fedbe} proposed FEDBE, a novel method to apply bayesian model ensemble into conventional federated learning, aiming at making the aggregation more robust. Motivated by prior work \cite{maddox2019simple}, the authors utilized bayesian inference to construct an improved global model. In addition, stochastic weight average (SWA) \cite{izmailov2018averaging} is also used to further boost the performance.

\subsection{Heterogeneous federated learning}
Heterogeneous federated learning aims to effectively aggregate models generated from heterogeneous environments. Here the heterogeneous property could be reflected from data, models or device systems. We will dive into each aspect in the next parts.

\subsubsection{Data heterogeneity}
Data heterogeneity indicates that collaborative clients might be in different situations, resulting in various data distributions. For example, the dog images collected from indoors and outdoors display highly heterogeneous data distribution. To address the issue, the research community borrows the idea from other AI techniques to alleviate the heterogeneity influence, which we list as follows.

\textit{Multi-task learning based methods.}
Multi-task learning enables learning models for multiple related tasks at the same time \cite{rebuffi2017learning,bilen2017universal,mallya2018packnet}. The core design principle is to capture the relationship among tasks and leverage the relationship to facilitate the learning  process. In federated learning, clients with different data distributions could also be considered as a type of multi-task learning, where each task has a distinctive statistical representation \cite{vanhaesebrouck2017decentralized,zantedeschi2020fully,t2020personalized,hanzely2020lower,hanzely2020federated,huang2021personalized}. 
For instance, Smith \textit{et al.} \cite{smith2017federated} first proposed to combine federated learning and multi-task learning. By a series of concept formulations and theoretical analyses, they suggested multi-task learning is a natural choice to handle the statistical problem in the federated setting. Based on the combination, they further developed a novel approach MOCHA, in order to accomplish their goal. Specifically, the authors formulated the problem as a dual optimization problem as follows
\begin{align}
\label{equation:mocha}
    \min _{\boldsymbol{\alpha}}\left\{\mathcal{D}(\boldsymbol{\alpha}):=\sum_{t=1}^{m} \sum_{i=1}^{n_{t}} \ell_{t}^{*}\left(-\boldsymbol{\alpha}_{t}^{i}\right)+\mathcal{R}^{*}(\mathbf{X} \boldsymbol{\alpha})\right\},
\end{align}
where $l_{t}^{*}$ and $\mathcal{R}^*$ are the conjugate dual functions of $l_{t}$ and $\mathcal{R}$, respectively. To solve \ref{equation:mocha}, they carefully designed the quadratic approximation of the dual problem to separate computation across the nodes. 

Despite federated multi-task learning being demonstrated effective, it has been applied only on convex models. To address the limitation, Corinzia \textit{et al.} \cite{corinzia2019variational} proposed a more general approach, named VIRTUAL, to achieve federation on non-convex models. 
The key idea is to construct a hierarchical Bayesian network in terms of the central server and the clients, such that the inference could be performed with variational methods. In this way, each client can obtain a task-specific model that benefits from the server model in a transfer learning manner.

Marfoq \textit{et al.} \cite{marfoq2021federated} further proposed to study federated multi-task learning under the flexible assumption that each local data distribution is a mixture of unknown underlying distributions, which is a more challenging and practical scenario. In the 
beginning, the authors showed the fact that t federated learning is impossible without
assumptions on local data distributions. Then they made the flexible assumption and developed Federated Expectation-Maximization to accomplish their objective. Besides, the proposed approach is proven generalizable to unseen clients.

\textit{Meta-learning based methods.}
Meta-learning is commonly considered as learning to learn \cite{thrun2012learning}. Compared with conventional deep learning algorithms that learn specific feature knowledge, meta-learning focus more on learning the learning ability. In the field of federated learning, meta-learning techniques can also be applied to generate a more personalized federation model. Jiang \textit{et al.} \cite{jiang2019improving} first proposed to combine them, where they believed meta-learning had a number of similarities with the objective of addressing the statistical challenge in FL. Concretely, they developed a novel algorithm to further combine FedAvg \cite{mcmahan2017communication} and Reptile \cite{nichol2018reptile}, with two modifications: the first one is to decrease the local learning rate to make training more stable; another is to design a fine-tuning stage based on Reptile with smaller K and Adam as the server optimizer, which could improve the initial model as well as preserving and stabilizing the personalized model. 


Khodak \textit{et al.} \cite{khodak2019adaptive} built a theoretical framework to further characterize meta-learning methods and apply them into federated learning. They introduced Average Regret-Upper-Bound Analysis (ARUBA), which enables meta-learning to leverage more sophisticated structures. With ARUBA, researchers could improve the results of many ML
 tasks, including adapting to the task-similarity, adapting to dynamic environments, adapting to the inter-task geometry and statistical learning-to-Learn. Towards FL, they improved meta-test-time performance on few-shot learning
and effectively added user-personalization to FedAvg. 

Fallah \textit{et al.} \cite{fallah2020personalized} aims to find an initial shared model that can be easily fitted to their local data with one or a few steps
of gradient descent. They achieved their objective by incorporating Model-Agnostic Meta-Learning (MAML) \cite{finn2017model,finn2018probabilistic} into current FL pipelines. Specifically, the authors proposed a personalized variant of the FedAvg algorithm, named Per-FedAvg, which can be formulated as optimizing the following equation
\begin{align}
\label{equation:Per-FedAvg}
    \min _{w \in \mathbb{R}^{d}} F(w):=\frac{1}{n} \sum_{i=1}^{n} f_{i}\left(w-\alpha \nabla f_{i}(w)\right),
\end{align}
where $n$ is the number of clients and $\alpha$ is the learning rate. The detailed solution for the optimization problem can be seen in the paper if readers have an interest.

\begin{figure}[t]
\centering
\includegraphics[width=1\columnwidth]{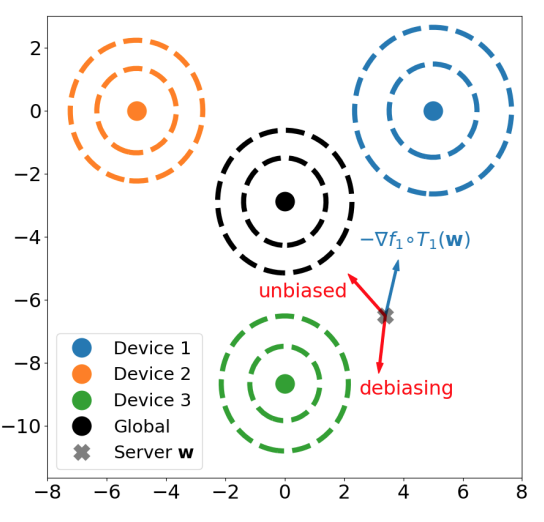} 
\caption{The illustration of PFL \cite{acar2021debiasing}.}
\label{fig:PFL}
\end{figure}

Acar \textit{et al.} \cite{acar2021debiasing} further modified meta-learning to benefit federated learning. As shown in Fig. \ref{fig:PFL}, they proposed PFL, a gradient correction method based on prior works, which explicitly de-biased the
meta-model in the distributed heterogeneous data setting to learn a more personalized device model. During the process, convergence guarantees of PFL for strongly convex, convex and nonconvex meta objectives are provided.

\textit{Transfer learning based methods.}
Transfer learning aims to transfer the information learned from a source task to a target task \cite{pan2009survey}. A large number of research works have been proposed to advance this promising field \cite{yosinski2014transferable,xuhong2018explicit,li2019delta,zhang2022remos}. In federated learning, transferring the knowledge of the federated model to each client model will significantly facilitate the personalization performance under the data heterogeneity environment. Wang \textit{et al.} \cite{wang2019federated} proposed to use fine-tuning, a typical transfer learning algorithm to achieve personalization. They first conducted traditional FL to obtain a global model. Then the federated model is regarded as the source model and further retrained using individual client’s training cache data. In this way, each client model can acquire and benefit the transferred knowledge, outputting an improved customized model.

Based on the aforementioned work, Yu \textit{et al.} \cite{yu2020salvaging} extended the simple fine-tuning strategy. They investigated how three adaptation mechanisms: fine-tuning,
multi-task learning, and knowledge distillation affect the personalization performance. The authors characterized these mechanisms as \textit{local adaptation}. In addition, different model protection techniques such as differential privacy and robust aggregation were applied to further validate the effectiveness of local adaptation. Finally, they used both CV and NLP datasets to demonstrate the superiority and necessity to conduct local adaptation.

Peng \textit{et al.} \cite{peng2019federated} considered a new FL+TL scenario beyond fine-tuning. Instead, they paid more attention to domain shift, which means that the labeled data collected by source nodes statistically differ from the target node’s unlabeled data. Based on this setting, they proposed the problem of federated domain adaptation and address it by Federated Adversarial Domain Adaptation (FADA). The key idea is to apply  adversarial adaptation and representation disentanglement to FL settings.

Ozkara \textit{et al.} \cite{ozkara2021quped} introduced a quantized and personalized FL algorithm to deal with the data issue. The quantized training process is conducted via knowledge distillation (KD) among clients who have access to heterogeneous data and resources. Besides, they developed an alternating proximal gradient update to address this compressed personalization challenge and analyzed its convergence properties.

\begin{figure}[t]
\centering
\includegraphics[width=1\columnwidth]{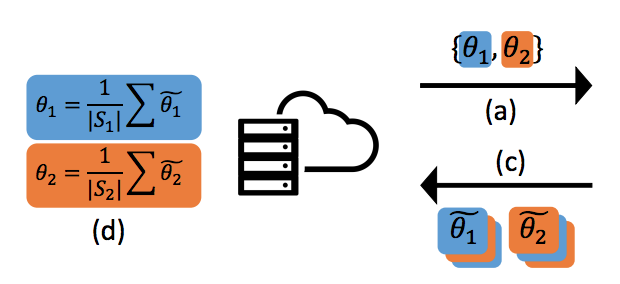} 
\caption{The illustration of IFCA \cite{ghosh2020efficient}.}
\label{fig:IFCA}
\end{figure}

\textit{Clustering-based methods.}
Clustering-based FL attempts to tackle the data heterogeneity issue via partitioning clients into different clusters, each of which conforms to a similar distribution. In terms of this key idea, much research effort is made to explore cluster-based FL. Sattler \textit{et al.} \cite{sattler2020clustered} proposed Clustered Federated Learning (CFL), to utilize geometric properties of the FL loss surface, in order to group
the client population into clusters with jointly trainable data distributions. It is worth noting that CFL is orthogonal to the current FL communication protocol and can be applied to
general non-convex objectives beyond DNNs.

Ghosh \textit{et al.} \cite{ghosh2020efficient} proposed the Iterative Federated Clustering Algorithm
(IFCA), which alternately estimated the cluster identities of the users and optimized model parameters for the user clusters via gradient descent. As shown in Fig. \ref{fig:IFCA}, the server broadcasted models and the workers dynamically identified their cluster memberships and run local updates. This process will continue to operate until the clusters become stable.

To train high-quality cluster models, Ruan \textit{et al.} \cite{ruan2022fedsoft} suggested FedSoft, which uses proximal updates to restrict client burden by asking a subset of clients to complete just one optimization task per communication round.


Liu \textit{et al.} \cite{liu2021pfa} proposed a framework to accomplish privacy-preserving federated adaptation. The key idea is to group the clients with similar distribution to collaboratively adapt the federated model, rather than just adapting it with the data in a single device. PFA leveraged the sparsity property of neural networks to generate privacy-preserving representations and used them to efficiently identify clients with similar data distributions. In this way, PFA can conduct an FL process in a group-wise way on the federated model to achieve adaptation.


Besides, in order to achieve clustering without uploading any extra information, Liu \textit{et al.} \cite{liu2021distfl} further proposed DistFL, targeting at finishing accurate, automated and efficient cluster-based FL in terms of distribution feature. Specifically, they extracted the distribution knowledge from the uploaded model via existing synthesis techniques \cite{mahendran2015understanding} and then compared them to obtain the clustering results. Finally, they aggregated models in each cluster, getting rid of the influence of heterogeneous data.

\subsubsection{Model heterogeneity}
Model heterogeneity means that the federated model might not be identical due to the different hardware and data distributions of clients. For example, in order to fit various
computation capabilities of clients, we require deploying different model architectures to match each client. On the other hand, NAS techniques \cite{zoph2016neural} have been widely used to search a crafted architecture based on the data in each device, thus leading to the model heterogeneity situation.

\begin{figure}[t]
\centering
\includegraphics[width=1\columnwidth]{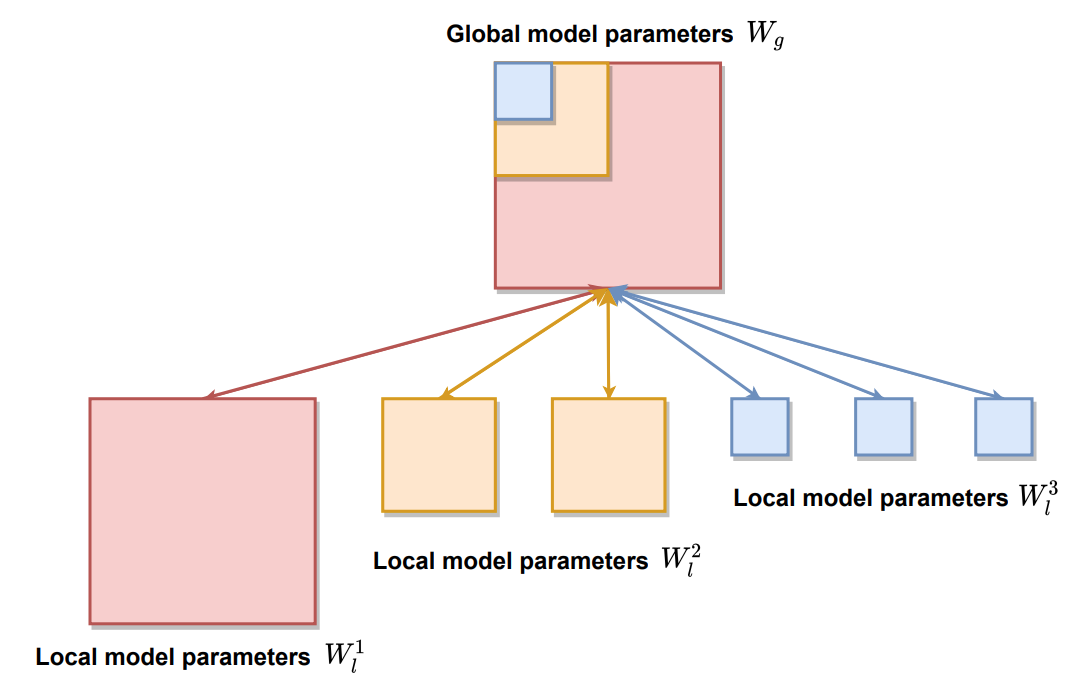} 
\caption{The illustration of HeteroFL \cite{diao2020heterofl}.}
\label{fig:heteroFL}
\end{figure}

To tackle the problem, Li \textit{et al.} \cite{li2019fedmd} used transfer learning
and knowledge distillation to develop a universal framework, which enabled federated
learning with uniquely designed models. Lin \textit{et al.} \cite{lin2020ensemble} further  proposed a distillation framework for robust federated model fusion and leveraged entropy-reduction to accelerate convergence. Diao \textit{et al.} \cite{diao2020heterofl} designed HeteroFL to address heterogeneous clients equipped with highly
different computation and communication capabilities. As shown in Fig. \ref{fig:heteroFL}, the federation is achieved by aggregating parameters on the same location while unlearning the other non-overlapping area.

\subsubsection{System heterogeneity}
System heterogeneity is a practical property in FL scenarios because different clients/parties naturally own heterogeneous hardware and memory limitation. Therefore, how to accomplish FL under the condition of system heterogeneity is worth exploring.

\begin{figure}[t]
\centering
\includegraphics[width=1\columnwidth]{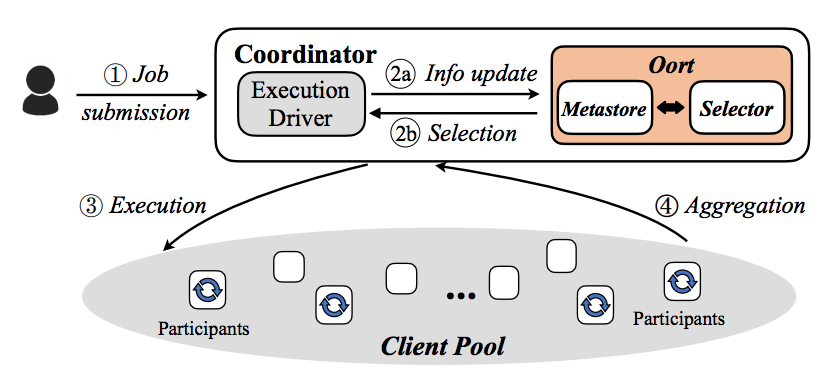} 
\caption{The illustration of Oort \cite{lai2021oort}.}
\label{fig:oort}
\end{figure}

A key design for system acceleration is to develop different client selection strategies for avoiding the influence of latency stragglers. Here stragglers refer to the clients with weak computing power and thus could slow down the overall FL process. Lai \textit{et al.} \cite{lai2021oort} proposed Oort, a system to improve the performance
of federated training and testing with guided participant selection. As shown in Fig. \ref{fig:oort}, Oort cherry-picked participants according to the tradeoff between statistical and system efficiency. Specifically, they defined "Client Statistical Utility" to measure the importance of each client. Shin \textit{et al.} \cite{shin2022fedbalancer} developed FedBalancer, a framework to actively select clients’ training samples in terms of the more ``informative" data. Besides, they introduced an adaptive deadline control scheme
to predict the optimal deadline for each round, in order to further speed up global training. Li \textit{et al.} \cite{lipyramidfl} observed that current client selection was coarse-grained due to their under-exploitation on the clients’ data
and system heterogeneity. Based on this finding, they proposed PyramidFL, a fine-grained client selection framework to  speed up the FL training. The key idea is to  not only focus on the divergence of those selected participants but also fully exploited the data and system heterogeneity within selected clients to profile their utility
more efficiently. As a result, PyramidFL is able to achieve better performance compared to other baselines.

\subsection{Secure federated learning}
The original design of federated learning considers the security problem via exchanging parameters while keeping raw data in their own devices. However, recent studies have proved that attackers might steal the privacy information from the uploaded models. Therefore, more rigorous secure FL should be investigated. In the following parts, we will introduce the attack methods and defense methods in FL scenarios.

\begin{figure}[t]
\centering
\includegraphics[width=1\columnwidth]{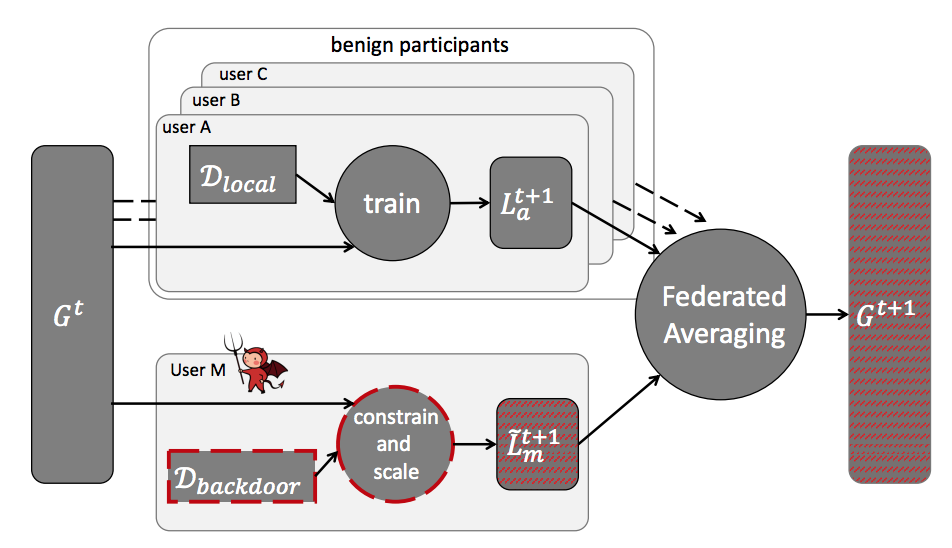} 
\caption{The illustration of model replacement \cite{bagdasaryan2020backdoor}.}
\label{fig:relpace}
\end{figure}

\subsubsection{Attack methods}
\textit{Backdoor attack.}
The goal of backdoor attacks is to manipulate a subset of training data by injecting adversarial triggers such that DNN models will
output incorrect prediction on the test set when the same trigger occurs. In federated learning, directly applying current backdoor attacks is unsuitable since the aggregation process might destroy the triggers. Bagdasaryan \textit{et al.}  \cite{bagdasaryan2020backdoor} is the first to backdoor federated learning. They achieved their objective by proposing model replacement, which means the backdoor is injected to the joint model rather than raw data. As shown in Fig. \ref{fig:relpace}, the attacker trained a model on the backdoor data using the constrain-and-scale technique. In this way, the averaging function is largely affected by this attack model. Wang \textit{et al.} \cite{wang2020attack} proposed  edge-case backdoors, which forced a model
to misclassify on seemingly easy inputs that are unlikely to be part of the
training or testing data. For example, they may exist on the tail of the input distribution. As a result, it is extremely hard to detect them. Xie \textit{et al.}  \cite{xie2019dba} further developed  distributed backdoor attack (DBA) to compromise FL. They mainly took advantage of the distributed nature of FL, decomposing a
global trigger pattern into separate local patterns and introducing them into the training set of different adversarial parties respectively. Therefore, DBA is more persistent and stealthy compared to centralized ones.
In FL models, backdoors can be inserted, but these backdoors are often not durable, i.e., they do not remain in the model after poisoned updates stop being uploaded. Since training occurs gradually in FL systems, an inserted backdoor may not survive until deployment. Zhang \textit{et al.}\cite{zhang2022neurotoxin} proposed Neurotoxin, which is a simple modification to existing backdoor attacks that target parameters that are not changed in magnitude as much during training.

\textit{Gradients attack.}
Gradients attack targets at reverse some privacy information from gradients. In federated learning, exchanging gradients is a typical step for knowledge update and aggregation. Therefore, gradient attack poses a high risk to the federal participants. Zhu \textit{et al.} \cite{zhu2019deep} found since training occurs gradually in FL systems, an inserted backdoor may not survive until deployment. that it is possible to obtain the
private training data from the publicly shared gradients. They first randomly generated a pair of “dummy” inputs and labels and used them to compute corresponding gradients. Then the gradients were compared to the shared ones and continually optimize the dummy inputs and labels to minimize the distance between them. As a result, the dummy data are close to the original ones and can peek into user privacy.
Lam \textit{et al.} \cite{lam2021gradient} further realized gradients attack from the aggregated model updates/gradients. The authors leveraged the summary information from device analytics and reconstructed the user participant matrix, which invalided the current secure aggregation protocols \cite{bonawitz2017practical}. 
Zhu \textit{et al.} \cite{zhu2020r} proposed Recursive Gradient Attack on Privacy (R-GAP), an approach to analyze how and when the target gradients can lead to the unique recovery of original data. Concretely, the authors designed a recursive, depth-wise algorithm for recovering training data from the gradient
information, which is the first closed-form algorithm that works on both CNN layers and FC layers. 
Li \textit{et al.} \cite{li2022auditing} found that under certain defense settings, generative gradient leakage can still leak private training information. 

\textit{Model poison attack.}
The goal of poison attacks is to induce the FL model to output the target label specified by the adversary. For example, Tolpegin \textit{et al.} \cite{tolpegin2020data} implemented data poison attack by flipping the labels of training data from one class to another class in the local training epoch to mislead the global model output. Although the aggregation process in FL can mitigate the attack to some extent, when the number of malicious clients becomes large, FL is inevitably poisoned. 
Fang \textit{et al.} \cite{fang2020local}  conducted the first systematic study
on local model poisoning attacks to federated learning. Based on this study, they proposed local model poisoning attacks to Byzantine robust federated learning via manipulating the
local model parameters on compromised worker devices during the learning process. Besides, the authors further stated two defense strategies and test their performance on the proposed attack.

\subsubsection{Defense methods}

\textit{DP-based defense.}
Differential privacy (DP) \cite{dwork2008differential} has been widely used to prevent information leakage. The key idea is to add some noises to obfuscate the original information. As a result, attackers are hard to infer the privacy properties. Federated learning also requires this type of protection since the uploaded model parameters can be easily exploited to extract sensitive information. 
Wei \textit{et al.} \cite{wei2020federated} proposed NbAFL, a framework that applied DP into FL. Specifically, they added noises to parameters of the local model at the client side before aggregation. Besides, the authors theoretically analyzed the convergence property of differentially private FL algorithms and proved the effectiveness of the proposed framework.

Kairouz \textit{et al.} \cite{kairouz2021distributed} 
presented a comprehensive end-to-end
system, where they discretized the data and added discrete Gaussian noise before conducting secure aggregation. In addition, the authors provided a novel privacy analysis for sums of discrete Gaussians and carefully analyzed the effects of data quantization and modular summation arithmetic. Experiments demonstrated that their method can achieve comparable performance with 16 bits of precision per value.
Agarwal \textit{et al.} \cite{agarwal2021skellam} proposed a  multi-dimensional Skellam mechanism, where two independent poisson random
variables are used to measure the difference. The authors applied their mechanism to FL and provided a novel algorithm that
appropriately discretized the data and used the Skellam mechanism along with modular arithmetic to bound the range of the data and communication costs before secure aggregation. As a result, they could achieve better  privacy-accuracy trade-offs in a more efficient manner.

\begin{figure}[t]
\centering
\includegraphics[width=1\columnwidth]{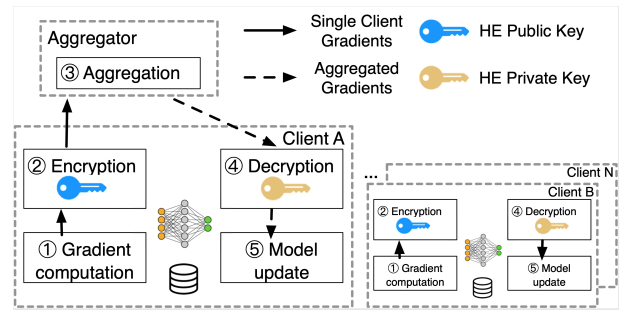} 
\caption{The illustration of BatchCrypt \cite{zhang2020batchcrypt}.}
\label{fig:EHE}
\end{figure}

\begin{figure*}[t]
\centering
\includegraphics[width=2\columnwidth]{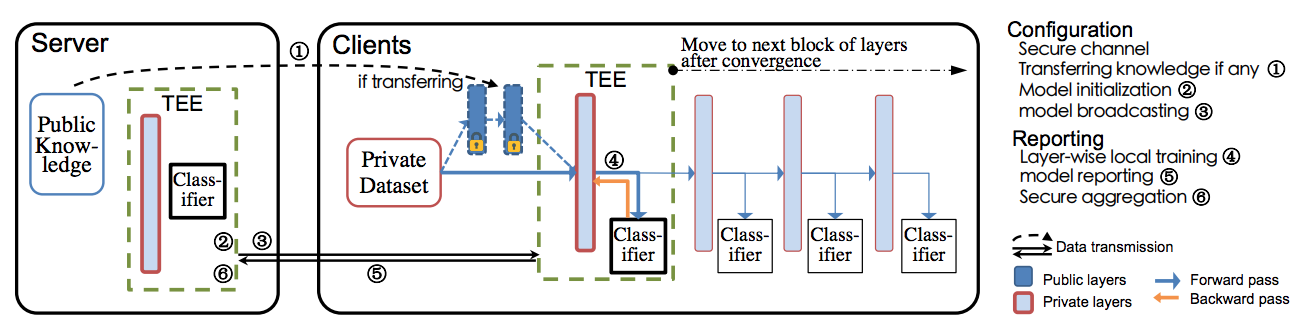} 
\caption{The illustration of PPFL \cite{mo2021ppfl}.}
\label{fig:PPFL}
\end{figure*}

\textit{HE-based defense.}
HE-based FL aims to combine traditional Homomorphic Encryption (HE) and FL in a more suitable way. By applying HE, FL is able to aggregate client models without revealing the information of the concrete model parameters. Therefore, it is impossible to infer user privacy from the model.
Hardy \textit{et al.} \cite{hardy2017private} proposed to encrypt FL with the homomorphic scheme in the field of privacy-preserving entity resolution and federated logistic regression. They bounded the difference between the empirical loss of their classifier on the true data and showed an improved convergence speed. Besides, their experiments found that even rates for generalization cannot be significantly affected by entity resolution.
Liu \textit{et al.} \cite{liu2019secure} designed a secure FL framework through leveraging the additive property of partial homomorphic encryption, which effectively avoids the exposure of client models at the server side. Besides, the authors introduced two optimization mechanisms to further enhance efficiency. 
Zhang \textit{et al.} \cite{zhang2020batchcrypt} proposed BatchCrypt, an efficient homomorphic encryption system for cross-Silo federated learning. As shown in Fig. \ref{fig:EHE}, there exist five typical steps to achieve a cross-silo FL system. In the beginning, the aggregator needed to select a client to generate an HE key-pair and distribute it to others. Then for each iteration, clients conducted local gradient updates and further encrypted them by the public key. These encrypted parameters were uploaded to the server where aggregation happened and the aggregated model is transferred to each client. Finally, the client side decrypted the received information and implemented the local training as the next round. BatchCrypt proposed two novel schemes to further improve efficiency. First, a feasible batch encryption scheme was presented to directly sum up the ciphertexts of two batches. Second, an efficient analytical model dACIQ was presented to choose optimal clipping thresholds with
the minimum cumulative error. As a result, BatchCrypt achieved 23×-93× training speedup while reducing the communication overhead by 66×-101×.

\textit{TEE-based defense.}
The aforementioned secure FL approaches provide security guarantee mainly from the perspective of software. In real-world scenarios, hardware protection is also widely applied by designing crafted architecture. Trusted Execution Environment (TEE) is 
a trusted component that establishes an isolated region on the main processor to ensure the confidentiality and integrity of data and programs \cite{arm2009security,costan2016intel}. Compared to traditional encryption schemes such as  homomorphic encryption, TEE is more efficient with respect to the computation cost since it only requires some simple operations to connect the trusted and untrusted part in OS. Recently there have been a large number of works targeting at applying TEE to deep/federated learning, in order to achieve protection from hardware level. For example, Mo \textit{et al.} \cite{mo2020darknetz} proposed DarkneTZ that enabled executing DNNs more secure with TEE in an edge device. They partitioned DNNs into a set of non-sensitive layers and sensitive layers, which are respectively processed by TEE or normal OS. Here the partition choice is based on the underlying system’s CPU execution time, memory usage, and accurate power consumption of different DNN layers. Besides, the authors developed a threat model to validate DarkneTZ's robustness under the membership inference attack and the results showed that DarkneTZ could defend against this type of attack with negligible performance overhead.

Based on the combination of DNNs and TEE, Mo \textit{et al.} \cite{mo2021ppfl} further attempted to apply TEEs to federated learning. Specifically, they proposed PPFL, a framework that limited privacy leakages in federated learning via implementing local training in TEEs. As shown in Fig. \ref{fig:PPFL}, to address the challenge of limited memory size of TEEs, the authors designed a greedy layer-wise training to conduct local updates until convergence. In this way, this approach could support sophisticated settings such as training one or more layers (block) each time, which potentially speed up the training process.
Zhang \textit{et al.} \cite{zhang2022teeslice} proposed TEESlice, a system to provide a strong security guarantee while maintaining low inference latency with the help of TEEs. Concretely, TEESlice executed the more private model slices on TEEs and others on normal AI accelerators. As a result, TEESlice can achieve more than 10× throughput promotion with the same level of strong security guarantee.

\subsection{Fair federated learning}

Existing works of federated learning pay more attention to improving learning performance based on the accuracy of the model and the time of learning task completion. However, the interests of the FL clients are often ignored and this may lead to unfairness. The problem of fairness can occur in the whole FL training process, including client selection, model optimization, incentive distribution, and contribution evaluation. The unfairness can have a negative impact on both the FL clients and the FL server, as clients are discouraged to join FL training, and servers are less likely to attract potentially high-quality clients. Recently, to achieve fairness from different angles, various Fairness-Aware Federated Learning (FAFL) approaches have been proposed. In this section, we will discuss recent FAFL methods in detail.

\subsubsection{Fair client selection}

Unfairness in FL Client Selection mainly consists of three types, over-representation, under-representation, and never-representation. Suppose an FL system prefers to select clients with high performance (such as a faster GPU), and clients with the highest performance may be selected much more than any other clients (i.e., over-representation), while clients with poor performance may be selected just a few times (i.e., under-representation). At the same time, the client with the lowest performance may never be selected (i.e., never-representation). Additionally, due to the heterogeneity among clients, fairness does not indicate giving everyone the same possibility to be selected. It is important to balance the interests of the server and the interests of the clients. If clients from specific groups are oversampled, the global FL model will be partial to their data, so the model's performance will deteriorate \cite{cho2020client}. Existing FAFL client selection methods can be partitioned into two categories, considering fairness factors and customization for each client. 

\textit{1) Fairness factors.}
Fairness factors are designed to allow rarely selected clients, such as clients with lower computational abilities or smaller datasets, to join the FL training more frequently. Yang \textit{et al.} \cite{yang2021federated} proposed a client selection algorithm based on the Combinatorial Multi-Armed Bandit (CMAB) framework to reduce the class imbalance effect. Inspired by \cite{li2019combinatorial}, Huang \textit{et al.} \cite{huang2020efficiency} converts the original offline problem to an online Lyapunov optimization problem and uses dynamic queues to quantify the long-term guarantee of the client participation rate. Moreover, Huang introduces a long-term fairness constraint to make sure the average client's long-term chosen rate is above a constant. After \cite{huang2020efficiency}, Huang \textit{et al.} \cite{huang2022stochastic} improves the performance by replacing dynamic queues to the Exp3 algorithms \cite{auer2002nonstochastic}, and the fairness parameter determining the selection possibility in each round can be different. However, these works all design the fairness factor without considering the real-time contribution of individual clients. Song \textit{et al.} \cite{song2021reputation} addresses this problem and proposed a client selection policy with fairness constraints based on reputation, using a fairness parameter to balance reputation and the number of successful transmissions. 

\textit{2) Client customization.}
This approach pays attention to customized model settings or customized model procedures. Clients often receive the same initial models at the first training round in most current FL paradigms. Therefore, clients with lower capabilities, such as bad network connections, require more time to complete each training round and are likely to be kept out of subsequent rounds, leading to under-presented and never-presented problems. To alleviate this problem, dynamically adapting the FL model framework or the training procedure based on client capabilities is often used. 

\begin{figure}[t]
\centering
\includegraphics[width=1\columnwidth]{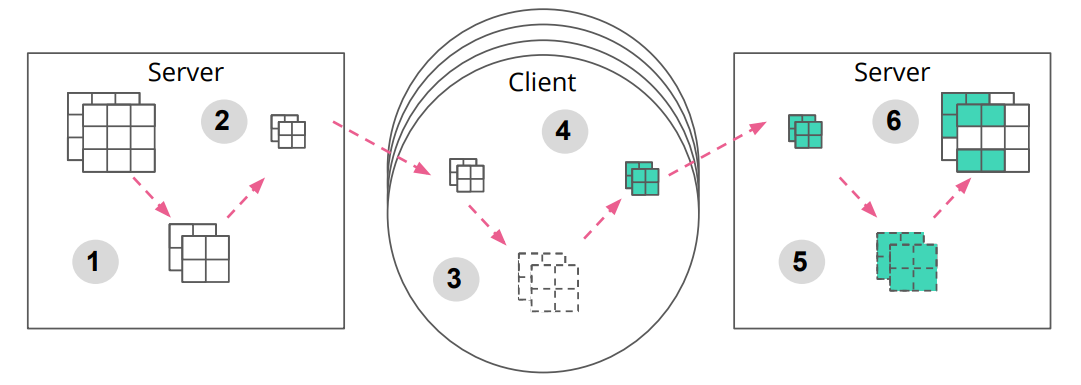} 
\caption{The summary of the Federated Dropout (FD) training procedure \cite{caldas2018expanding}. }
\label{fig:FDmodel}
\end{figure}

Caldas \textit{et al.} \cite{caldas2018expanding} proposed Federated Dropout (FD), which distributes sub-models with sizes suitable for each client based on their computational resources. The process of FD is shown in Fig \ref{fig:FDmodel}. Although FD diminishes communication and local computation costs largely, it uses dropout operations and treats the neural networks as black-box functions. Bouaciada \textit{et al.} noticed this problem and proposed Adaptive Federated Dropout (AFD) \cite{bouacida2021adaptive}. AFD keeps an activation score map to generate the best-fit sub-model for each client. FD and AFD both make sure clients with low capabilities could participate in FL training, but they do not provide custom pruned submodels to different clients. To address this limitation, Horvath \textit{et al. } \cite{horvath2021fjord} augmented FD to Ordered Dropout (OD). Different from FD, OD drops neighboring components of the model despite random neurons. OD divides clients with comparable computational capabilities into clusters, and clients in the same cluster apply the same dropout rate. Moreover, OD applies the knowledge distillation method \cite{hinton2015distilling} to enhance feature extraction for smaller submodels. 

\begin{figure}[t]
\centering
\includegraphics[width=1\columnwidth]{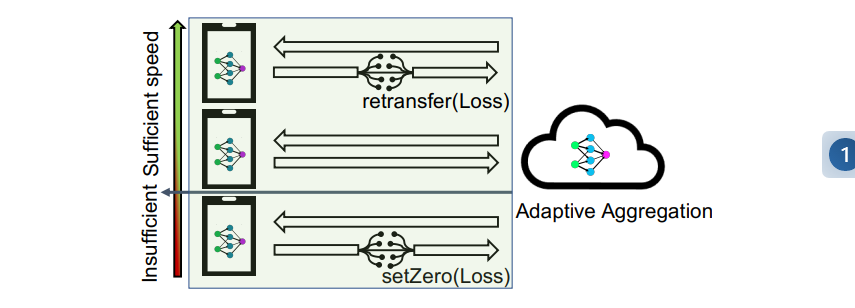} 
\caption{The illustration of ThrowRightAway (TRA) scheme \cite{zhou2021loss}.}
\label{fig:TRAmethod}
\end{figure}


Clients' communication capabilities can also affect client selection. A poor network may cause too much retransmission and lead to extra delays in FL model training, which makes clients with a poorer network less likely to aggregate their model updates into the final model and leads to model bias. To deal with this issue, Zhou \textit{et al.} \cite{zhou2021loss} proposed ThrowRightAway (TRA), a loss-tolerant FL framework that makes the FL training faster by ignoring few lost packets. As is shown in Fig \ref{fig:TRAmethod}, at first every participating FL client reports their network conditions to the FL server, and the server divides the clients into two categories: sufficient type and insufficient type. Only the clients in the sufficient type can get a re-transmission request and then re-transmit their loss packets. Apparently, the method can only be effective when the category is accurate. 


This method means assigning less work to clients with lower capabilities to make them available to pass threshold-based FL client selection. Li \textit{et al.} \cite{li2020federated} proposed FedProx which allowed each client performed partial training based on its accessible resources. FedProx allows various local epochs, and thus more clients are encouraged to join the training process. 

\subsubsection{Fair model optimization}

In the optimization during FL model training, the model may discriminate against definite preserved groups, or overfit some clients at the expense of others. Recent works dealing with this issue can be approximately divided into two types: 1) objective function-based and 2) gradient-based. 

\textit{1) Objective function-based methods: }
Objective function-based methods focus on the global/local objectives of the FL model, such as minimizing the loss function. Mohri \textit{et al.} \cite{mohri2019agnostic} proposed AFL, which aims to prevent the model overfitting any specific client at the expense of others. AFL just optimizes the global model for the target distribution made up of a mixture of clients. However, this method only works for a small number of clients. Zhou \textit{et al.} \cite{zhou2021loss} proposed q-FFL to diminish the scalability limitation of AFL. q-FFL adds parameter q to reweigh the aggregate loss. To improve the model robustness and maintain good-intent fairness at the same time, Hu \textit{et al.} \cite{hu2022federated} proposed fedMGDA+ which optimizes each FL client's loss function respectively and simultaneously. Addressing the same issue, Li \textit{et al.} \cite{li2021ditto} proposed Ditto, which improves fairness and robustness at the same time. 

While the methods mentioned all pay attention to the accuracy parity notion of fairness, there are also many kinds of research focusing on group fairness. Du \textit{et al.} \cite{du2021fairness} proposed AgnosticFair, which incorporates an agnostic fairness constraint. Although it has good accuracy and fairness on unknown testing data distribution, it needs prior knowledge to design the re-weighting function, which limits its application in dynamic systems. Cui \textit{et al.} \cite{cui2021addressing} proposed FCFL, a multi-objective optimization framework that achieves good-intent fairness and group fairness at the same time. Different from AFL, it minimizes the loss of the client with the worst performance and uses a smooth surrogate maximum function considering all clients. A fairness constraint is also added to calculate the disparities among all clients. 

\textit{2) Gradient-based approaches: }
Here, gradient means the local updated gradient of each client in every local iteration. Wang \textit{et al.} \cite{wang2021federated} proposed the federated fair averaging (FedFV) algorithm, which aims to average clients' gradients after mitigating potential conflicts among clients. FedFV detects gradient conflicts through the cosine similarity and modifies both the direction and magnitude of the gradients by iteratively eliminating such conflicts. However, the estimated gradients may be incompatible with the latest updates. 

\subsubsection{Fair contribution evaluation}
Contribution evaluation in FL learning indicates that an FL system can evaluate the contribution of different clients without accessing data from the clients. Many methods designed for non-privacy machine learning environments cannot be applied to FL scenarios directly. A general method is to evaluate each client's model contribution to the aggregated FL model, and a fair evaluation is critical. Unfairness in contribution evaluation may lead to the free-rider issue \cite{hardin2003free}, which implies that clients contribute little but can get similar benefits as the clients who contribute more. In this part we will introduce five types of existing FL contribution evaluation methods with their typical works.

\begin{figure}[t]
\centering
\includegraphics[width=1\columnwidth]{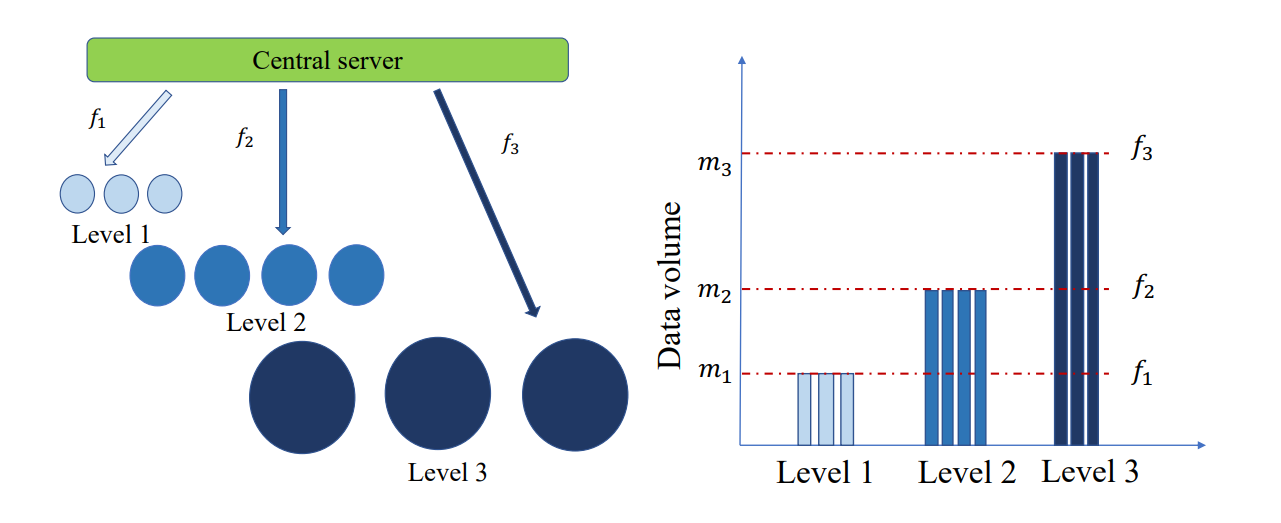} 
\caption{The illustration of hierarchically fair federated learning (HFFL) \cite{zhang2020hierarchically}. }
\label{fig:HFFL}
\end{figure}

\textit{1) Self-reported information: }
This method of evaluation contribution is based on clients reporting their information actively. Most works based on this method believe their clients are reliable, which is not always correct in practice. Proposed by Zhang \textit{et al.} \cite{zhang2020hierarchically}, Hierarchically fair federated learning (HFFL) follows the idea of 'contribute more, get more reward', which is proved effective in social psychology \cite{tornblom1985subrules}, game theory \cite{rabin1993incorporating} and bandwidth allocation \cite{li2008proportional}. Hence, it's critical to figure out how to evaluate a client's contribution and how much proportion of reward a client should get to ensure fairness. Data Shapley can be used to evaluate contribution in machine learning, but Shapley value is model-dependent \cite{ghorbani2019data} and incompatible with FL tasks. As a result, Zhang proposes evaluating contributions based on publicly verifiable factors of clients, such as cost of data collection, data volume, and data quality, to avoid the inconsistency of model-dependent methods. To distribute proportional rewards to clients, Zhang introduces hierarchically fair federated learning (HFFL), as is shown in Figure \ref{fig:HFFL}. The publicly verifiable factors determined by the clients' consensus about each client are reported to the FL server, and the FL server then uses the information to rate each client, which at the same level are supposed to contribute to the model equally and will get the equal reward. 

\textit{2) Individual evaluation: }
Individual evaluation implies evaluating contribution through performance on specific tasks and pays more attention to individual performance instead of global performance. The method often adopts two assumptions that both the server and the client are reliable and clients with a similar model to others are regarded to supply more contribution, which is not always feasible. To achieve fairness without sacrificing the model performance, Lyu \textit{et al.} \cite{lyu2020collaborative} proposed a Collaborative Fair Federated Learning (CFFL) framework based on reputation, which uses a reputation mechanism to achieve collaborative fairness. Lyu definites collaborative fairness as the reward is proportional to the client's contribution. Different standard FL process, CFFL allows clients to receive only the allocated aggregated updates according to their reputations, and the server is in charge of a reputation list which is updated in each communication round relying on the quality of the uploaded gradients of each participant. 

\textit{3) Utility game: }
The utility game \cite{gollapudi2017profit} refers to a game where each player chooses an available team to maximize their payoffs, while the universal social welfare is the total utility produced by all the teams cumulatively. FL contribution evaluation methods based on utility games have a deep connection with profit-sharing schemes, and there are three diffusely used profit-sharing schemes: (1) Egalitarian: any part of the utility produced by a team is separated equally between the members.
(2) Marginal gain: the payoff of a player in a team is equal to the team gained when the player joined.
(3) Marginal loss: the payoff of a player in a team is equal to the team will lose if the player leaves.

Among the three types above, the marginal loss scheme is the most commonly adopted. Wang \textit{et al.} \cite{wang2019measure} proposed a deletion method to evaluate contributions in horizontal federated learning. This evaluation method consists of removing the instances supplied from one definite party, retraining the model, calculating the difference between the original model and the new model, and using this difference to define the contribution of this party. Wang formulates the influence measure as follows, 
\begin{equation}
Influence^{-i} = \frac{1}{n}\sum_{j=1}^{n}\left | \hat{y}_{j} - \hat{y}_{j}^{-i} \right |,
\end{equation}
where n is the size of the dataset, $\hat{y}_{j}$ is the model trained on all data prediction on \textit{j}th instance, and $\hat{y}_{j}^{-i}$ s the model trained without the \textit{i}th instance prediction on \textit{j}th instance.

Then Huang defines a party's contribution as the total influence of all instances it possesses.
\begin{equation}
Influence^{-D} = \sum_{i\in D}Influence^{-i},
\end{equation}

For vertical horizontal learning, Huang uses shapley value which will be introduced in the next part.

\textit{4) Shapley value: }
Shapley value (SV) was first introduced in cooperative game theory \cite{shapley1997value}. Different from marginal loss, SV-based FL contribution evaluation approaches can reflect the contribution of a client's own data, in spite of its joining order, and can produce a fairer evaluation. However, SV's computational complexity is $O(2^n)$, so many approaches have been proposed to improve efficiency. 


\section{Prevalent frameworks of federated learning}
\label{sec:framework}


In this section, we will introduce several prevalent frameworks of federated learning, including FedLab, Flower, FedML, FATE, and FedScale.

\textbf{FedLab.}
Since most FL schemes follow the same basic steps and just a few changes in some steps are needed in different scenarios, Zeng \textit{et al.} \cite{zeng2021fedlab} proposed FedLab \cite{SMILELab-FL}, which is designed flexible and customizable, offers essential functional modules, and has highly customizable interfaces. Two main roles in FL settings are provided: Server and Client, and both of them are made up of two components, NetworkManager and ParameterServerHandler/Trainer. The design focuses more on communication efficiency and FL algorithm effectiveness. To support methods improving Communication Efficiency, FedLab uses tensor-based communication, supports customizable communication agreement, and implements both Synchronous and Asynchronous communication patterns according to Federated Optimization algorithms. For Optimization Effectiveness, FedLab applies a "high-cohesion and low-coupling" optimization module which provides aggregation and data partition methods. Additionally, FedLab can be used in various scenarios, such as Standalone, Cross-process and Hierarchical FL simulation.

\textbf{Flower.}
Due to the lack of frameworks that are able to support scalably executing FL methods on mobile and edge devices, Beutel \textit{et al.} \cite{beutel2020flower} proposed Flower \cite{FLOWER}, which can run large-scale FL experiments on different FL device scenarios. Flower makes it possible to smoothly transition from experimental research to system research on a large group of real edge devices. Designed to be scalable, client-agnostic, communication-agnostic, privacy-agnostic, and flexible, Flower has extensive implementations, such as communication stack, serialization, ClientProxy, and Virtual Client Engine(VCE).

\textbf{FedML.}
Proposed by He \textit{et al.} \cite{he2020fedml}, FedML \cite{FedML-AI} aims to solve the lack of support for diverse FL computing paradigms, support of diverse FL configurations, and standardized FL algorithm implementations and benchmarks. FedML library is mainly made up of high-level API FedML-API and low-level API FedML-core. To support FL on real-world hardware platforms, FedML offers on-device FL testbeds called FedML-Mobile and FedML-IoT which are built upon real-world hardware platforms. FedML programming interface allows worker/client-oriented programming, message definition beyond gradient and model, topology management, trainer and coordinator, privacy, security, and robustness, so users can just pay attention to algorithms implementations and ignore the backend details. 

\textbf{FATE.}
Since most open-sourced frameworks are research-oriented and lack the implementation on industry, Liu \textit{et al.} \cite{liu2021fate} proposed FATE(Federated AI Technology Enabler) \cite{webank_fl}, which is the first production-oriented platform. Built on FederatedML, FATE provides Private Set Intersection(PSI), and uses distributed computation framework Eggroll to improve computation efficiency. FATE provides three main components, scheduling system FATE-Flow, visualization tool FATE-Board, and high-performance inference platform FATE-Serving. In addition, kinds of deployments are supported, including building FATE on top of Kubernetes in data centers through KubeFATE, manual or docker deployments on Mac and Linux, and cross-cloud deployment and management through FATE-cloud.

\textbf{FedScale.}
Lai \textit{et al.} \cite{lai2021fedscale} proposed FedScale \cite{FedScale}, which contains many realistic FL datasets for different tasks, and FedScale Runtime which is an automated evaluation platform aiming to simplify and standardize FL evaluation in more realistic environments. The raw data of FedScale datasets are collected from various sources, processed into consistent formats, sorted into different FL use cases and packed into standardized APIs for users to easily use in other frameworks. The evaluation platform, FedScale Runtime, is equipped with both mobile and cluster backends to enable both on-device FL evaluation on smartphones, and FL evaluations in real deployments and in-cluster simulations. 

\section{Discussion}
\label{sec:discussion}
This section summarizes some limitations of current FL approaches and discusses possible future directions.

\textbf{Dynamic federated learning.}
Current federated learning approaches assume that data in each client are stable and unchanged. However, in real-world scenarios, clients may be in an ever-changing environment, where the local data are continuously observed and processed by sensors. Under this condition, directly conducting conventional training and aggregation will suffer from the catastrophic forgetting problem, which indicates that the prior knowledge learned by the model might be forgotten as new data arrive. Incremental learning \cite{castro2018end,wu2019large,lomonaco2021avalanche} is a hot research topic to address the issue, targeting at learning new knowledge
while maintaining the ability to recognize previous ones. In the future, how to effectively combine federated learning and incremental learning is worth exploring.

\textbf{Decentralized federated learning.}
A central server is of vital importance to traditional federated learning since aggregation needs to be conducted in this side. Considering that the third-party server may not be honest, uploading parameters or gradients to it potentially exists security risks. Therefore, it is necessary to achieve federated learning without a server involved.
Although He \textit{et al.} \cite{he2019central} has made a preliminary attempt to decentralized FL, they only target logistic regression and the experiments are insufficient.
How to accomplish general decentralized FL still remains an open problem.

\textbf{Scalability of federated learning.}
Recent FL papers paid more attention to designing new algorithms to improve FL performance under different conditions. However, they ignore the scalability property, which determines whether we could operate large-scale FL. In many cooperation scenarios, there might be a huge number of parties and we should provide guidance to the cooperation improvement as the number of participants increases. In a word, FL scalability deserves future investigation.

\textbf{Unified benchmark.}
Although a large number of datasets have been used for evaluating the performance of FL, there is still a lack of a unified benchmark to align the results for a fair comparison. On one hand, in order to achieve different federated goals (e.g., personalization, robustness), researchers use different datasets to test the performance. On the other hand, two typical types of FL, horizontal FL and vertical FL, also apply distinctive datasets to demonstrate the performance of different FL types. Thus a unified benchmark will definitely benefit the FL community.

\section{Conclusion}
\label{sec:conclusion}
Federated learning has gained more and more attention due to its ability of collaboratively generating a global model without leaking sensitive information. Recent surveys have summarized many related works devoted to offering a comprehensive understanding to developers and readers in this community. However, most of them focus on a specific aspect of FL or fail to catch the latest progress of this hot research topic. This paper provides a systematic survey, which investigates recent development on federated learning. By analyzing the pipeline and challenges of FL, we propose a taxonomy with different FL aspects involved. In addition, we also explore some practical FL frameworks and characterize their features. Finally, some limitations and future direction are concluded in order to promote the evolution of the FL community.

\balance
\bibliographystyle{ACM-Reference-Format}
\bibliography{sample-base}
\clearpage

\end{document}